\documentclass{article}

\usepackage{amsmath, amsthm, hyperref}

\title{Practical Boolean Backpropagation}
\author{Simon Golbert}
\date{\today}

\theoremstyle{definition}
\newtheorem*{definition}{Definition}

\theoremstyle{remark}
\newtheorem*{note}{Note}

\begin{document}

\maketitle

\begin{abstract}
    Boolean neural networks offer hardware-efficient alternatives to real-valued models. While quantization is common, purely Boolean training remains underexplored. We present a practical method for purely Boolean backpropagation for networks based on a single specific gate we chose, operating directly in Boolean algebra involving no numerics. Initial experiments confirm its feasibility.
\end{abstract}

\section{Introduction}
To reduce the computational complexity and memory requirements of models, various quantization techniques are widely used today. Floating-point numbers are typically reduced to 16-, 8-, or 4-bit representations. This technique allows preserving the traditional gradient-based optimization of a differentiable loss function through backpropagation.

More recent research is moving toward more extreme forms of quantization. For instance, \cite{ma2024} introduces BitNet b1.58, a novel 1.58-bit Large Language Model (LLM) where each parameter is represented using ternary values \{-1, 0, 1\}. Another prominent study, \cite{rastegari2016}, presents XNOR-Net, a method for efficient image classification using purely Boolean convolutional neural networks. Despite achieving very decent inference efficiency, the latter project (like the former) relies on training driven by traditional backpropagation (albeit with heavily quantized parameter values), thus still requiring significant computational resources to run.

To maximize efficiency in both model training and inference, we should consider a methodology based on purely Boolean computations. This approach has the potential to leverage existing hardware for massive parallelism while also opening avenues for developing specialized, cost-effective, and energy-efficient hardware optimized for bitwise operations. While not yet a definitive solution, exploring this direction could lead to significant advancements in high-performance computing.

In this article, we will define a composite Boolean gate to serve as a neuron in Boolean artificial neural networks (ANNs). Additionally, we will derive an error backpropagation routine to enable training. Finally, we will discuss practical aspects of the proof-of-concept (PoC) implementation and share key observations from our experiments.

\section{Model Structure}
We define the gate function, the related \textbf{Row Activation} operation, and the inference process for a simple model composed of fully connected layers.

\subsection{Gate Function}
This research is focused on utilizing the following composite gate as a neuron function:
\[ y = \bigvee_{i=1}^{n} \left( x_{i} \wedge w_{i} \right) \oplus b \]
Here \( x_{i} \) are the argument values, \( w_{i} \) and \( b \) are parameter values that are learned during training.

\begin{note}
    To save space, from now on, we will use 1 instead of True and 0 instead of False to represent the Boolean values throughout the document.
\end{note}

It's easy to see that the gate above is complete. Indeed we can express the OR, AND, and NOT gates using this gate:
\[ \text{NOT}(x) = \left( x \wedge 1 \right) \oplus 1 \]
\[ \text{OR}(x_1, x_2) = \left( \left( x_1 \wedge 1 \right) \vee \left( x_2 \wedge 1 \right) \right) \oplus 0 \]
\[ \text{AND}(x_1, x_2) = \left( \left( \text{NOT}(x_1) \wedge 1 \right) \vee \left( \text{NOT}(x_2) \wedge 1 \right) \right) \oplus 1 \]

\subsection{Row Activation}
In order to express model inference in matrix form using the gate function defined above, we introduce the following operation.

\begin{definition}
    \textbf{Row Activation} is an operation that is defined for two Boolean matrices \( X \) and \( W \) of size \( m \times n \), denoted as:
    \[ Z = \mathcal{A}(X, W) \]
    and produces a \( 1 \times m \) Boolean row vector \( Z \). Each element \( Z_{1i} \) (for \( i = 1, 2, \dots, m \)) is computed as:
    \[ Z_{1i} = \bigvee_{j=1}^{n} \left( X_{ij} \wedge W_{ij} \right). \]
    In other words, \( Z_{1i} \) is true if there exists at least one column \( j \) such that both \( X_{ij} \) and \( W_{ij} \) are true.

    The \textbf{Row Activation} operation can also be applied when one of the arguments (either \( X \) or \( W \)) is a \( 1 \times n \) Boolean row vector, while the other remains an \( m \times n \) matrix. In this case, the vector is broadcast to an \( m \times n \) matrix by repeating it across all \( m \) rows before applying.
\end{definition}

\subsection{Fully Connected Layer}
Now we can define a fully connected layer as \( (W, B) \), where \( W \) is a matrix of "weights" and \( B \) is a vector of "biases". For the input vector \( X \), the output of the layer is computed as:
\[ Y = \mathcal{A}(X, W) \oplus B \]
Here, \( B \) is a \( 1 \times m \) Boolean row vector that selectively inverts elements of the \textbf{Row Activation} output via elementwise XOR.

A simple model can be represented as a series of such layers, where during inference, the output of the \( i \)-th layer serves as the input to the \( (i+1) \)-th layer.

\section{Model Training}
In this section, we introduce key definitions that will aid in outlining a training process, which conceptually resembles traditional error backpropagation.

\subsection{Activation Sensitivity}
To direct the training process, we need to determine which elements contribute to the final result and which do not. This concept is similar to the gradient in traditional training, but instead of quantifying influence, Boolean values indicate whether a change in an element flips the final output.

For instance, consider the following \textbf{Row Activation} application:
\[
    X = \begin{bmatrix} 1 & 0 & 1 & 0 \end{bmatrix}, \quad
    W = \begin{bmatrix} 0 & 0 & 0 & 0 \\ 0 & 1 & 0 & 1 \\ 1 & 0 & 1 & 0 \end{bmatrix}, \quad
    Z = \mathcal{A}(X, W) = \begin{bmatrix} 0 & 0 & 1 \end{bmatrix}
\]
Now, examine which elements of \( X \) need to be flipped to change each element of the result. It is evident that flipping any element of \( X \) cannot affect \( Z_{1,1} \), since \( W_1 \) is zeroed. Flipping \( X_2 \) or \( X_4 \) can change \( Z_2 \) from 0 to 1. To change \( Z_3 \) from 1 to 0, both \( X_1 \) and \( X_3 \) must be flipped.

An important observation that can be distilled from the example above is that flipping a resulting element from 0 to 1 requires flipping \textbf{any} of the relevant argument elements, while flipping a resulting element from 1 to 0 requires flipping \textbf{all} of the relevant arguments.

Now we can define operations that determine such "sensitive" elements.

\begin{definition}
    For given Boolean matrices \( A \) and \( B \) of size \( m \times n \), and \( Z = \mathcal{A}(A, B) \), the \textbf{Positive Activation Sensitivity} operation, denoted as
    \[ S = \mathcal{S}^+(A, B), \]
    produces a matrix \( S \) of the same size, where each element \( S_{i,j} \) is defined as:
    \[ S_{i,j} =
        \begin{cases}
            1, & \text{if } Z_{1,i} = 0 \text{ and setting } A_{i,j} = 1 \text{ causes } Z_{1,i} \text{ to flip to } 1, \\
            0, & \text{otherwise}.
        \end{cases}
    \]

    The \textbf{Negative Activation Sensitivity} operation, denoted as
    \[ S = \mathcal{S}^-(A, B), \]
    produces a matrix \( S \) of the same size, where each element \( S_{i,j} \) is defined as:
    \[ S_{i,j} =
        \begin{cases}
            1, & \text{if } Z_{1,i} = 1 \text{ and setting } A_{i,j} = 0 \text{ is necessary for } Z_{1,i} \text{ to flip to } 0, \\
            0, & \text{otherwise}.
        \end{cases}
    \]

    The \textbf{Activation Sensitivity} operation is defined as
    \[ \mathcal{S}(A, B) = \mathcal{S}^+(A, B) \vee \mathcal{S}^-(A, B) \]
    The operations above can also be applied when one of the arguments (either \( A \) or \( B \)) is a \( 1 \times n \) Boolean row vector, while the other remains an \( m \times n \) matrix. In this case, the vector is broadcast to an \( m \times n \) matrix by repeating it across all \( m \) rows before applying.
\end{definition}

In the example above:
\[
    \mathcal{S}(X, W) = \begin{bmatrix} 0 & 0 & 0 & 0 \\ 0 & 1 & 0 & 1 \\ 1 & 0 & 1 & 0 \end{bmatrix}, \quad
    \mathcal{S}(W, X) = \begin{bmatrix} 1 & 0 & 1 & 0 \\ 1 & 0 & 1 & 0 \\ 1 & 0 & 1 & 0 \end{bmatrix},
\]
It's easy to see that for a fully connected layer \( \mathcal{S}(X, W) \) and \( \mathcal{S}(W, X) \) define the sensitivity of \( Y \) to changes in \( X \) or \( W \), respectively, regardless of \( B \). Indeed, flipping \( z \) always flips \( z \oplus b \) regardless of the value of \( b \).

\subsection{Error Projection}
Before formally defining the \textbf{Error Projection} operation, we first develop the intuitive logic that justifies its existence.

Consider a single fully connected layer trained on Boolean input vectors \( X_1, X_2, \dots, X_b \) of size \( 1 \times n \) and their corresponding expected Boolean output vectors \( Y^e_1, Y^e_2, \dots, Y^e_b \) of size \( 1 \times m \). For the given input vectors, the layer produces the corresponding output vectors \( Y_1, Y_2, \dots, Y_b \). We then compute the output errors as \( E_1, E_2, \dots, E_b \), where \( E_k = Y_k \oplus Y^e_k \).

Our aim is to find \( D^w \) that minimizes the total Hamming weight (i.e., the number of 1s) in all the errors for subsequent inference after applying \( W' = W \oplus D^w \). Since each row \( W_i \) is processed independently of the others during inference, this task can clearly be reduced to a single row.

For a given \( i \), assume we have all minimal difference masks \( D_k \) for each \( k \) that flip \( Y_{k,1,i} \) when using \( X_{k,1,i} \oplus D_k \) instead of \( X_{k,1,i} \). Now, let us construct two matrices by concatenating \( D_k \) as rows: \( C \) for indices where \( E_{k,1,i} = 0 \) and \( I \) for output vectors where \( E_{k,1,i} = 1 \). To construct a proper \( D^w_i \), we need to apply as many \( I \)-rows as possible to flip incorrect outputs while ensuring that \( C \)-rows are not activated, as their activation "spoils" correct outputs.

Here is a concrete example. Assume we have:
\[
    W = \begin{bmatrix}
        1 & 0 & 0 & 1 & 0 & 1 \\
        0 & 1 & 1 & 0 & 1 & 0 \\
        0 & 0 & 1 & 0 & 0 & 1
    \end{bmatrix}, \quad
    B = \begin{bmatrix} 0 & 1 & 0 \end{bmatrix}
\]
\[
    \begin{aligned}
        X_1 & = \begin{bmatrix} 1 & 1 & 0 & 1 & 0 & 1 \end{bmatrix}, \\
        X_2 & = \begin{bmatrix} 0 & 0 & 1 & 0 & 0 & 0 \end{bmatrix}, \\
        X_3 & = \begin{bmatrix} 0 & 0 & 0 & 0 & 0 & 1 \end{bmatrix},
    \end{aligned} \quad
    \begin{aligned}
        Y^e_1 & = \begin{bmatrix} 1 & 1 & 1 \end{bmatrix} \\
        Y^e_2 & = \begin{bmatrix} 1 & 0 & 0 \end{bmatrix} \\
        Y^e_3 & = \begin{bmatrix} 1 & 1 & 0 \end{bmatrix}
    \end{aligned}
\]
First, we compute outputs and errors:
\[
    \begin{aligned}
        Y_1 & = \begin{bmatrix} 1 & 0 & 0 \end{bmatrix}, \\
        Y_2 & = \begin{bmatrix} 0 & 0 & 1 \end{bmatrix}, \\
        Y_3 & = \begin{bmatrix} 1 & 1 & 1 \end{bmatrix},
    \end{aligned} \quad
    \begin{aligned}
        E_1 & = \begin{bmatrix} 0 & 1 & 1 \end{bmatrix} \\
        E_2 & = \begin{bmatrix} 1 & 0 & 1 \end{bmatrix} \\
        E_3 & = \begin{bmatrix} 0 & 0 & 1 \end{bmatrix}
    \end{aligned}
\]
Now we select the first element of the outputs to derive \( D^w_1 \) without loss of generality, as the remaining \( D^w_2 \) and \( D^w_3 \) can be obtained in the same manner.

It is easy to see that the minimal difference masks can be derived by selecting individual bits from the rows of the \textbf{Positive Activation Sensitivity} matrix or by taking entire rows from the \textbf{Negative Activation Sensitivity} matrix, depending on the corresponding \textbf{Row Activation} output values.

\[
    C = \begin{bmatrix}
        1 & 0 & 0 & 1 & 0 & 1 \\
        0 & 0 & 0 & 0 & 0 & 1
    \end{bmatrix} \quad
    I = \begin{bmatrix}
        1 & 0 & 0 & 0 & 0 & 0 \\
        0 & 0 & 0 & 1 & 0 & 0 \\
        0 & 0 & 0 & 0 & 0 & 1
    \end{bmatrix}
\]
The matrix \( C \) above contains difference mask rows that "spoil" \( Y_{1,1,1} \) and \( Y_{3,1,1} \), while \( I \) consists of masks, each designed to "fix" \( Y_{2,1,1} \). It is clear that \( I_1 \) and \( I_2 \) do not "spoil" the outputs even when combined, but \( I_3 \) conflicts with \( C_2 \), and when combined with the former two, it also conflicts with \( C_1 \), so it should definitely be discarded. As a result, we get:
\[
    D^w_1 = \begin{bmatrix} 1 & 0 & 0 & 1 & 0 & 0 \end{bmatrix}
\]

After we have obtained the necessary intuition, we can formally define the \textbf{Error Projection} operation.

\begin{definition}
    \textbf{Error Projection} is an operation that is defined for two Boolean matrices, \( C \) of size \( p \times n \), and \( I \) of size \( q \times n \), denoted as:
    \[ D = \mathcal{R}(C, I) \]
    and produces a \( 1 \times n \) Boolean row vector,
    \[ D = \bigvee_{i=1}^{h} I'_{i}, \]
    where \( I' \) is the largest possible subset of the rows of \( I \) such that
    \[ \forall i, \quad C_i \wedge D \neq C_i. \]
\end{definition}

\section{Error Backpropagation}
Before gradually deriving an error backpropagation routine, we introduce an auxiliary operation that facilitates a smooth transition from the \textbf{Activation Sensitivity} matrix to difference masks, which serve as input for \textbf{Error Projection}.

\begin{definition}
    \textbf{Selection Expansion} is an operation defined for a Boolean vector \( S \) of size \( 1 \times n \), denoted as:
    \[ M = \mathcal{E}(S) \]
    It produces a \( p \times n \) Boolean matrix \( M \), where \( p \) is the number of 1 elements in \( S \). The matrix \( M \) is constructed by vertically concatenating rows, where for each \( S_{1,i} = 1 \), there is a corresponding row in \( M \) with a 1 in the \( i \)-th position and 0s elsewhere.
\end{definition}

The initial step of the \textbf{Error Backpropagation} routine is to derive a difference mask \( D^w \) such that substituting \( W \) with \( W' \) minimizes the output errors, where
\[ W' = W \oplus D^w \]

First, we compute the \textbf{Activation Sensitivity} matrices, which represent the sensitivity of \textbf{Row Activation} to changes in \( W \) when applied to \( X_k \):
\[ S^{w+}_k = \mathcal{S^+}(W, X_k), \quad S^{w-}_k = \mathcal{S^-}(W, X_k) \]

Second, each row of \( D^w_i \) is computed as:
\[ D^w_i = \mathcal{R}(C_i, I_i) \]
where
\[
    C^+_i = \begin{bmatrix} \vdots \\ \mathcal{E}(S^{w+}_{k,i}) \\ \vdots \end{bmatrix}_{Z_{k,i} = 0} \quad
    C^-_i = \begin{bmatrix} \vdots \\ S^{w-}_{k,i} \\ \vdots \end{bmatrix}_{Z_{k,i} = 1} \quad
    C_i = \begin{bmatrix} C^+_i \\ C^-_i \end{bmatrix}
\]
for \( k \) such that \( E_{k,i} = 0 \), and
\[
    I^+_i = \begin{bmatrix} \vdots \\ \mathcal{E}(S^{w+}_{k,i}) \\ \vdots \end{bmatrix}_{Z_{k,i} = 0} \quad
    I^-_i = \begin{bmatrix} \vdots \\ S^{w-}_{k,i} \\ \vdots \end{bmatrix}_{Z_{k,i} = 1} \quad
    I_i = \begin{bmatrix} I^+_i \\ I^-_i \end{bmatrix}
\]
for \( k \) such that \( E_{k,i} = 1 \).

\begin{note}
    The \(\begin{bmatrix} \quad \end{bmatrix}\) notation here and below designates vertical rank-preserving matrix concatenation. For example, for an \( m \times n \) matrix \( A \) and a \( p \times n \) matrix \( B \), the matrix \( C = \begin{bmatrix} A \\ B \end{bmatrix} \) is of size \( (m+p) \times n \) and consists of the rows from \( A \) and \( B \).
\end{note}

Using the updated \( W' \), we compute \( Z'_k \), \( Y'_k \) for each \( k \), and the corresponding error \( E'_k \) as follows:
\[ Z'_k = \mathcal{A}(X_k, W'), \quad Y'_k = Z'_k \oplus B, \quad E'_k = Y'_k \oplus Y^e_k \]

To eliminate common errors where \( E_{k,i} = 1 \) for all \( k \), we modify the bias \( B \) using a conjunction over all error vectors:
\[ D^b = \bigwedge_{k=1}^b E'_k, \quad B' = B \oplus D^b \]

The errors remaining after applying the corrections with \( W' \) and \( B' \) are defined as \( E''_k \), which must be corrected by earlier layers:
\[ E''_k = E'_k \wedge \overline{D^b} \]

To correct the errors, we need to find the optimal difference masks \( D^x_k \) for each \( X_k \) that minimize the number of 1s in the corresponding \( E''_k \). These difference masks will then serve as the errors \( E_k \) for the preceding layer.

First, we compute the \textbf{Activation Sensitivity} matrices, which represent the sensitivity of \textbf{Row Activation} to changes in \( X_k \) when applied to \( W' \):
\[ S^{x+}_k = \mathcal{S^+}(X_k, W'), \quad S^{x-}_k = \mathcal{S^-}(X_k, W') \]

The matrices above allow us to obtain minimal difference masks that flip each \( i \)-th element in \( Y''_k = Z'_k \oplus B' \). Similarly to the computation of \( D^w_i \), we need to identify as many difference masks as possible that correct the output values where \( E''_{k,i} = 1 \), while preserving the values where \( E''_{k,i} = 0 \).

Next, we compute \( D^x_k \) by reusing the \textbf{Error Projection} operation, as described in the previous step:
\[ D^x_k = \mathcal{R}(C_k, I_k) \]
where
\[
    C^+_k = \begin{bmatrix} \vdots \\ \mathcal{E}(S^{x+}_{k,i}) \\ \vdots \end{bmatrix}_{Z'_{k,i} = 0} \quad
    C^-_k = \begin{bmatrix} \vdots \\ S^{x-}_{k,i} \\ \vdots \end{bmatrix}_{Z'_{k,i} = 1} \quad
    C_k = \begin{bmatrix} C^+_k \\ C^-_k \end{bmatrix}
\]
for \( i \) such that \( E''_{k,i} = 0 \), and
\[
    I^+_k = \begin{bmatrix} \vdots \\ \mathcal{E}(S^{x+}_{k,i}) \\ \vdots \end{bmatrix}_{Z'_{k,i} = 0} \quad
    I^-_k = \begin{bmatrix} \vdots \\ S^{x-}_{k,i} \\ \vdots \end{bmatrix}_{Z'_{k,i} = 1} \quad
    I_k = \begin{bmatrix} I^+_k \\ I^-_k \end{bmatrix}
\]
for \( i \) such that \( E''_{k,i} = 1 \).

We have just described the \textbf{Error Backpropagation} iteration for the final fully connected layer. The preceding layers follow the same routine, applied sequentially from the penultimate layer back to the first. The only difference is that, instead of computing the initial errors as \( E_k = Y_k \oplus Y^e_k \), these layers receive their errors from the desired input difference masks of the subsequent layer: \( E_k = D^x_k \).

\section{Specialization}
The \textbf{Error Backpropagation} routine described above follows a single flow. It relies on the \textbf{Error Projection} operation, which has relatively high computational complexity. This could potentially negate the performance advantages of a purely Boolean training approach compared to traditional methods. In fact, optimizing this operation remains an open question and requires further research, which is beyond the scope of this article.

We can significantly reduce the complexity of the \textbf{Error Projection} by ignoring cases where the output flips only when multiple input elements change simultaneously—that is, when rows in the \textbf{Negative Activation Sensitivity} matrices contain more than one 1.

\begin{definition}
    For given Boolean matrices \( A \) and \( B \) of size \( m \times n \), and \( Z = \mathcal{A}(A, B) \), the \textbf{Specialized Activation Sensitivity} operation, denoted as
    \[ S = \mathcal{S}^*(A, B), \]
    produces a matrix \( S \) of the same size, where each element \( S_{i,j} \) is defined as:
    \[ S_{i,j} =
        \begin{cases}
            1, & \text{if flipping } A_{i,j} \text{ alone causes } Z_{1,i} \text{ to flip}, \\
            0, & \text{otherwise}.
        \end{cases}
    \]
    Similarly to \textbf{Row Activation}, the \textbf{Specialized Activation Sensitivity} operation can be applied when one of the arguments (either \( A \) or \( B \)) is a \( 1 \times n \) Boolean row vector, while the other remains an \( m \times n \) matrix. In this case, as before, the vector is broadcast to an \( m \times n \) matrix by repeating it across all \( m \) rows before applying.
\end{definition}

With the \textbf{Specialized Activation Sensitivity}, there is no longer a need for \textbf{Selection Expansion}, as the corresponding \textbf{Specialized Error Projection} can now be computed efficiently using simple element-wise Boolean operations. Specifically, given the condensed matrices \( C \) and \( I \) — constructed from rows of the \textbf{Specialized Activation Sensitivity} matrices, where each 1 indicates that flipping a single element suffices to change the output — we can easily determine an optimal difference mask. This is done by selecting combinations that fix the errors while simply excluding those that would "spoil" correct outputs.

\begin{definition}
    \textbf{Specialized Error Projection} is an operation defined for two Boolean matrices \( C \) of size \( p \times n \), and \( I \) of size \( q \times n \), denoted as:
    \[ D = \mathcal{R}^*(C, I), \]
    which produces a \( 1 \times n \) Boolean row vector \( D \):
    \[ D_{1,j} = \left( \bigvee_{i=1}^{q} I_{i,j} \right) \wedge \neg \left( \bigvee_{i=1}^{p} C_{i,j} \right) \quad \text{for each } j = 1 \text{ to } n.
    \]
\end{definition}

\begin{note}
    In practice, when a \textbf{Specialized Error Projection} vector contains multiple 1s, it is essential to zero all but one of them to prevent the zeroing of rows in subsequent \textbf{Specialized Activation Sensitivity} matrices, which will negate further learning. A random selection of the 1 to retain is likely the most effective approach.
\end{note}

Now, we derive the \textbf{Error Backpropagation} steps for the specialized case, focusing only on the differences from the general case.

First, to derive \( D^w \), we compute the \textbf{Specialized Activation Sensitivity} matrices, which represent the sensitivity of \textbf{Row Activation} to changes in \( W \) when applied to \( X_k \):
\[ S^{w*}_k = \mathcal{S}^*(W, X_k) \]

Next, we compute each row \( D^w_i \) of the difference mask \( D^w \) as:
\[ D^w_i = \mathcal{R}^*(C_i, I_i) \]
where
\[ C_i = \begin{bmatrix} \vdots \\ S^{w*}_{k,i} \\ \vdots \end{bmatrix}_{E_{k,i} = 0} \quad I_i = \begin{bmatrix} \vdots \\ S^{w*}_{k,i} \\ \vdots \end{bmatrix}_{E_{k,i} = 1} \]

Then, to derive \( D^x_k \), we compute the \textbf{Specialized Activation Sensitivity} matrices, which represent the sensitivity of \textbf{Row Activation} to changes in \( X_k \) when applied to \( W' \):
\[ S^{x*}_k = \mathcal{S}^*(X_k, W') \]

Finally, we compute the difference mask \( D^x_k \) for each \( k \) as:
\[ D^x_k = \mathcal{R}^*(C_k, I_k) \]
where
\[ C_k = \begin{bmatrix} \vdots \\ S^{x*}_{k,i} \\ \vdots \end{bmatrix}_{E_{k,i} = 0} \quad I_k = \begin{bmatrix} \vdots \\ S^{x*}_{k,i} \\ \vdots \end{bmatrix}_{E_{k,i} = 1} \]

While the specialized routine seems to offer reduced learning capabilities, a thorough comparison with the general approach requires further research, which is beyond the scope of this article.

\section{Discussion}
Our proposed Boolean backpropagation method demonstrates a feasible approach to training neural networks using purely Boolean operations. Compared to traditional methods like those in \cite{rastegari2016}, which rely on numerical gradients even with quantized weights, our method eliminates floating-point computations entirely, potentially reducing computational overhead.

Initial experiments with the specialized approach suggest that the model can learn complex patterns. For instance, a model with 4 fully connected layers with widths \( 6272 \rightarrow 4096 \rightarrow 4096 \rightarrow 4096 \rightarrow 320 \) is capable of recognizing MNIST digits with 75\% accuracy after 30 minutes of training on a laptop CPU \cite{bbp-research}. These results indicate that, despite its limitations, the specialized approach can be competitive with traditional backpropagation in certain tasks.

\section{Conclusion}
While the proposed Boolean backpropagation approach shows promise, further investigation is needed to fully understand its limitations and potential applications. Future work will focus on refining the method to handle more complex tasks and determining its scalability across different network architectures. Additionally, a thorough comparison with other state-of-the-art approaches will be crucial in identifying areas for improvement.

\end{document}